\pdfoutput=1 
\documentclass[a4paper, 10pt, conference]{ieeeconf}      
\IEEEoverridecommandlockouts                              
\usepackage[english]{babel}
\usepackage{blindtext}
\usepackage[framemethod=tikz]{mdframed}
\usepackage{stfloats}
\usepackage{graphics} 
\usepackage{epsfig} 
\usepackage{times} 
\usepackage{amsmath} 
\usepackage{amssymb}  
\usepackage{cite}
\usepackage{multirow}
\usepackage{multicol}
\usepackage{adjustbox}
\usepackage{graphicx}

\usepackage{hyperref}
\usepackage{url}
\usepackage{subfig}
\usepackage{booktabs}
\usepackage{array}
\usepackage{arydshln}
\usepackage[framemethod=tikz]{mdframed}
\usepackage{longtable}
\usepackage{xtab,afterpage}
\usepackage{diagbox}
\usepackage{makecell}
\usepackage{slashbox,pict2e}
\usepackage{romannum}
\usepackage{verbatim}
\usepackage{algorithm}
\usepackage{algpseudocode}
\usepackage{siunitx}
\usepackage{textcomp}
\usepackage{placeins}
\usepackage{mwe}
\usepackage[english]{babel}
\usepackage{blindtext}
\usepackage{breqn}
\usepackage{stfloats}
\usepackage{mathdots}
\usepackage{esvect}
\usepackage{caption}
\usepackage{kotex}

\newcommand{\rom}[1]{\uppercase\expandafter{\romannumeral #1\relax}}

\title{\LARGE \bf
PaGO-LOAM: Robust Ground-Optimized LiDAR Odometry
}

\author{Dong-Uk Seo$^{1*}$, Hyungtae Lim$^{1*}$, \textit{Student Member, IEEE},\\ Seungjae Lee$^{1}$, \textit{Student Member, IEEE}, and Hyun Myung$^{1}$, \textit{Senior Member, IEEE}
\thanks{$*$These authors contributed equally to this work}
\thanks{$^{1}$Donguk Seo, $^{1}$Hyungtae Lim, $^{1}$Seungjae Lee, and $^{1}$Hyun Myung are with the School of Electrical Engineering and KI-AI at KAIST (Korea Advanced Institute of Science and Technology), Daejeon, 34141, South Korea. {\tt\small \{dongukseo, shapelim, sj98lee, hmyung\}@kaist.ac.kr}}
\thanks{This work was supported by the Industry Core Technology Development Project, 20005062, Development of Artificial Intelligence Robot Autonomous Navigation Technology for Agile Movement in Crowded Space, funded by the Ministry of Trade, Industry \& Energy (MOTIE, Republic of Korea). The students are supported by the BK21 FOUR from the Ministry of Education (Republic of Korea).}
}

\begin{document}
\maketitle
\thispagestyle{empty}
\pagestyle{empty}
\begin{abstract}
Numerous researchers have conducted studies to achieve fast and robust ground-optimized LiDAR odometry methods for terrestrial mobile platforms. In particular, ground-optimized LiDAR odometry usually employs ground segmentation as a preprocessing method. This is because most of the points in a 3D point cloud captured by a 3D LiDAR sensor on a terrestrial platform are from the ground. However, the effect of the performance of ground segmentation on LiDAR odometry is still not closely examined. In this paper, a robust ground-optimized LiDAR odometry framework is proposed to facilitate the study to check the effect of ground segmentation on LiDAR SLAM based on the state-of-the-art (SOTA) method. By using our proposed odometry framework, it is easy and straightforward to test whether ground segmentation algorithms help extract well-described features and thus improve SLAM performance. In addition, by leveraging the SOTA ground segmentation method called \textit{Patchwork}, which shows robust ground segmentation even in complex and uneven urban environments with little performance perturbation, a novel ground-optimized LiDAR odometry is proposed, called \textit{PaGO-LOAM}. The methods were tested using the KITTI odometry dataset. \textit{PaGO-LOAM} shows robust and accurate performance compared with the baseline method. Our code is available at \footnotesize\href{https://github.com/url-kaist/AlterGround-LeGO-LOAM}{\texttt{https://github.com/url-kaist/AlterGround-LeGO-LOAM}\normalsize}
\end{abstract}


\section{Introduction} \label{sec:intro}

In recent years, Simultaneous Localization and Mapping~(SLAM) has been studied due to its numerous applications on various mobile platforms such as autonomous driving~\cite{shan2018lego, sung2021whatif, geiger2012kitticvpr, behley2019semantickitti}, unmanned aerial vehicles (UAVs)~\cite{lee2021real}, unmanned ground vehicles (UGVs)~\cite{scaramuzza20111dransac}, mobile phones~\cite{lim2021avoiding}, and so forth. Of course, SLAM can be performed by using various sensors, e.g. RGB or RGB-D cameras~\cite{song2022g2p}, radar sensors, ultrasonic sensors. Some researchers have employed a 3D light detection and ranging (LiDAR) sensor to achieve a precise 3D point cloud map~\cite{song21flooprplan, lim2020normal}. 3D LiDAR sensors are relatively expensive compared with other sensors but they can acquire long-range measurements, as well as provide centimeter-level accuracy. These advantages allow LiDAR sensors to be used for various tasks, for instance, localization~\cite{kim2018rgb}, point cloud registration~\cite{segal2009gicp, lim2022quatro, kim2019gp}, mapping~\cite{lim2020msdpn,javanmardi2017towards}, and so forth.

 In this paper, we \textcolor{black}{specifically} focus on LiDAR odometry frameworks for terrestrial platforms, and there are many odometry frameworks using a 3D LiDAR sensor~\cite{zhang2014loam,pan2021mulls,jung2020bridge,xu2021fast}, which show outstanding pose accuracy and scalability. In particular, some LiDAR odometry frameworks leverages the fact that terrestrial mobile platforms come into contact with ground~\cite{lim21erasor}, so ground-aided odometry methods also have been proposed~\cite{shan2018lego,kim2018scancontext,wei2021groundslam}. 
 
In the meanwhile, numerous researchers have conducted studies to achieve a fast and robust ground segmentation~\cite{lim2021patchwork,zhang2015ground,mongus2014ground,rummelhard2017ground}. Ground segmentation can be utilized as a preprocessing method because most of the points in a 3D point cloud captured by a 3D LiDAR on a terrestrial platform are from the ground in outdoor environments. In addition, ground are usually flat, so ground segmentation also helps extract planar features in surroundings roughly. For this reason, ground segmentation is already implemented in LeGO-LOAM~\cite{shan2018lego}. However, the effect of the performance of ground segmentation on LiDAR odometry is still not closely examined. In other words, in~\cite{shan2018lego}, the performance with and without the application of ground constraints was not studied. 

\begin{figure}[t!]
    \captionsetup{font=footnotesize}
	\centering 
	\includegraphics[width=0.48\textwidth]{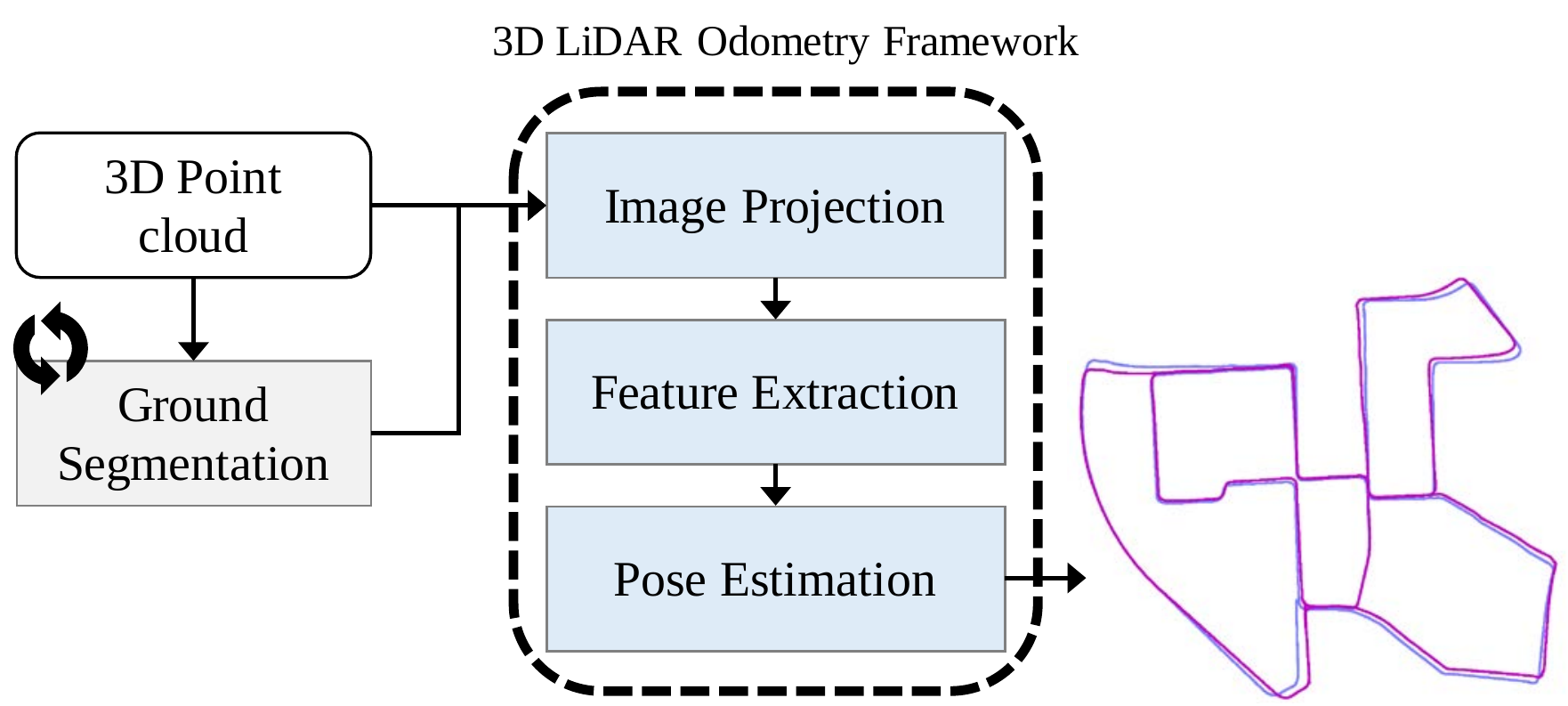}
	\caption{Overview of our ground segmentation-based LiDAR odometry framework for terrestrial platforms to study the effect of ground segmentation on LiDAR odometry. By using our ground segmentation-agnostic framework, one can test the effect of ground segmentation on LiDAR odomety easily. Accordingly, ground segmentation module can be changed very easily.}
	\label{fig:overview}
\vspace{-0.5cm}
\end{figure}
 
Therefore, a robust ground-optimized LiDAR odometry framework is proposed to facilitate the study to check the effect of ground segmentation on LiDAR odomety based on the state-of-the-art~(SOTA) method~\cite{shan2018lego}, as shown in Fig.~\ref{fig:overview}. By using our odometry framework, it is easy and simple to test whether ground segmentation algorithms help extract well-described features and thus improve SLAM performance. In addition, by leveraging the SOTA ground segmentation method called \textit{Patchwork}~\cite{lim2021patchwork}, which shows robust ground segmentation even in complex, uneven urban environments with little performance perturbation, a novel ground-optimized LiDAR odometry is proposed, called \textit{PaGO-LOAM}.

In summary, the contribution of this paper is threefold: 
\begin{itemize}
	\item To the best of our knowledge, it is the first open source to analyze the impact of the ground segmentation method on LiDAR odometry easily in complex urban environments. 
	\item In addition, a robust ground-optimized LiDAR odometry framework is also proposed by using our previous work~\cite{lim2021patchwork}.
	\item Our proposed method shows a promising performance compared with the baseline method~\cite{shan2018lego} in both with and without the loop-closure. In particular, our method shows significant pose accuracy in rural environments where ground is more uneven and bumpy. 
\end{itemize}

\section{Related Works} \label{sec:rw}

\subsection{Ground Segmentation}

Several researchers have studied ground segmentation methods~\cite{na2016drivable, byun2015drivable,zermas2017fast,himmelsbach2010fast,narksri2018slope,lim2021patchwork,cheng2020simple}. Ground segmentation is typically utilized for two purposes. One is to find the traversable area \cite{na2016drivable, byun2015drivable} in navigation, and the other is to segment a point cloud to achieve object tracking or recognition~\cite{himmelsbach2010fast,narksri2018slope,lim2021patchwork}. In particular, the latter case is based on the fact that terrestrial objects, such as vehicles or humans, come into contact with the ground~\cite{lim21erasor}. Accordingly, once the ground is removed successfully given a 3D point cloud, objects can be classified by simple clustering methods, such as Euclidean clustering. Furthermore, 
because many point clouds belong to the ground, once ground segmentation is performed as preprocessing, computational cost can be greatly reduced when detecting the objects~\cite{cheng2020simple}.  


\subsection{Point Cloud Registration}

In order to obtain odometry from dense point clouds, 3D point cloud registration, which estimates a relative pose through the most appropriate matching between points, was proposed. Accordingly, when sequential point clouds come in, odometry could be calculated by accumulating relative poses according to time sequence. A representative one is Iterative Closest Point (ICP)~\cite{besl1992method}, and it has made an impact on subsequent studies. Unfortunately, the ICP-variants set point pairs by using a greedy, exhaustive nearest neighbor~(NN) search for every iteration, so they are only applicable when two point clouds are close enough or {nearly} overlapped~\cite{pomerleau2013comparing}. Otherwise, the correspondences are likely to become invalid. Under the circumstance, the result of the registration may get caught in the local minima~\cite{koide2020vgicp,lim2022quatro}.
In addition, using all the points as correspondences requires a lot of computational costs; thus, this necessarily leads to the appearance of fast and lightweight relative pose estimation methods, i.e. feature-based methods.

\subsection{LiDAR Odometry Framework}

Odometry framework usually requires to work in real-time, so the feature-based method has been proposed~\cite{zhang2014loam,shan2018lego,shan2020lio}. One of renowned methods is LOAM~\cite{zhang2014loam}. In LOAM, edge features and planar features are extracted respectively to estimate relative motion between two consecutive frames. However, LOAM does not discern ground points, which make up the majority within a point cloud, and non-ground points that potentially results in increase of computational cost. In LeGO-LOAM~\cite{shan2018lego}, an advanced method of LOAM~\cite{zhang2014loam}, ground segmentation is introduced to get more granular features. 
By creating and utilizing the two range images, one for upcoming point cloud and the other for the ground, edge and planar features were extracted. Unfortunately, their ground segmentation method is a line-based method, so it is sensitive to some noises atypical objects such as lawns or bushes. In~\cite{shan2020lio}, LIO-SAM was proposed whose pose estimation accuracy was improved by using preintegration of an inertial measurement unit sensor. Nevertheless, directly processing the edges and planar features without ground might also be a similar disadvantage as ~\cite{zhang2014loam}. 
In ~\cite{pan2021mulls}, various features (line, edge, surface, etc.) are extracted, and matching correspondences between them are used for the estimation. Since \cite{pan2021mulls} uses the ground as a feature different from the surface, there is also a possibility to be improved by changing the ground extraction method.
Nevertheless, as \cite{shan2018lego} has better usability to separate the module for evaluating the improvement of changing the ground segmentation method, we end up selecting \cite{shan2018lego} as the baseline code for creating the ground module.


\begin{figure*}[h]
    \captionsetup{font=footnotesize}
	\centering 
	\includegraphics[width=16cm]{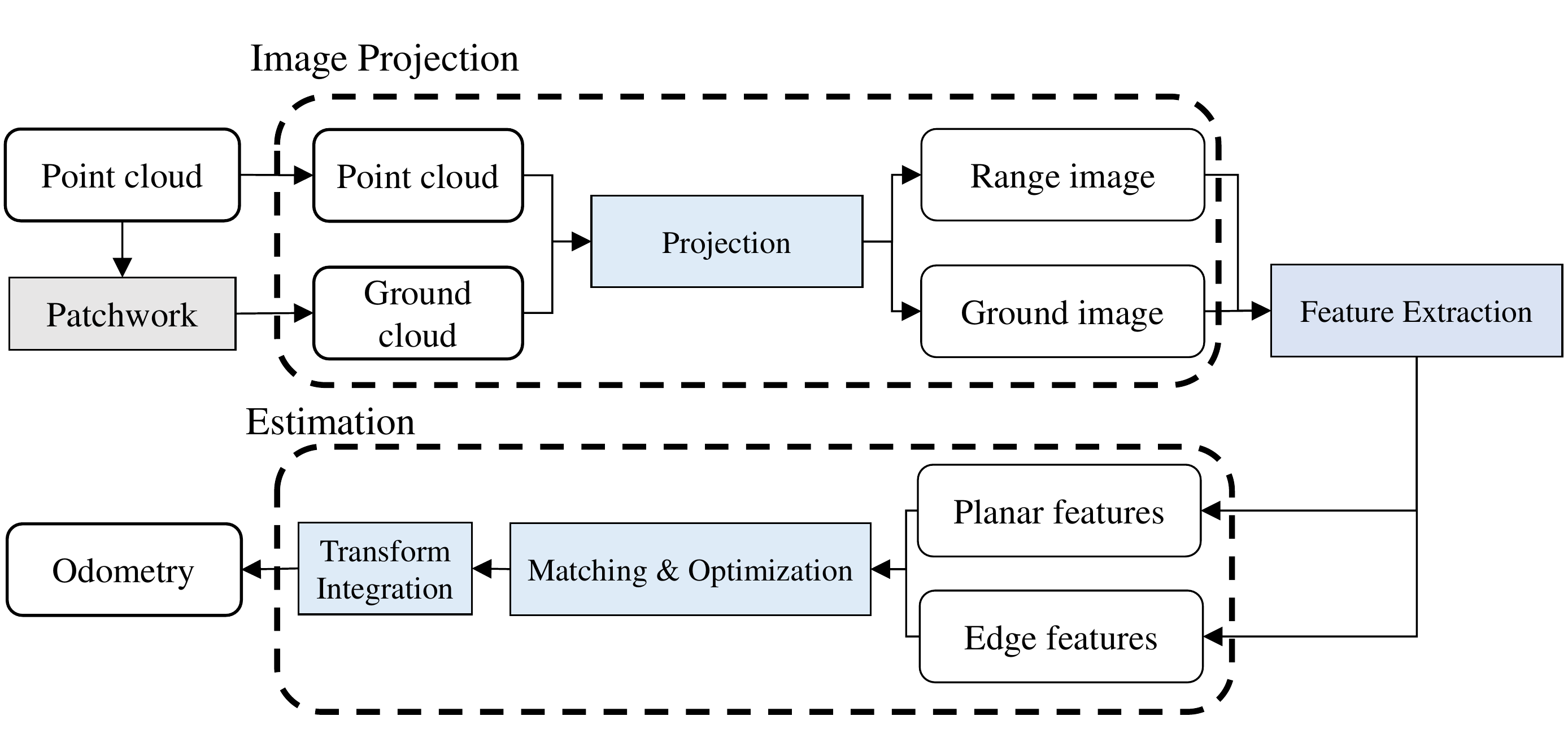}
	\caption{The detailed process of our ground-optimized LiDAR odometry framework. In this figure, Patchwork~\cite{lim2021patchwork} is employed as an example of utilization of our LiDAR framework.}
	\label{fig_flow}
\end{figure*}


\subsection{Ground-Constrained LiDAR Odometry}

As an extension of LeGO-LOAM~\cite{shan2018lego}, there are some approaches to employ ground segmentation in their odometry frameworks. 
For instance, Ground SLAM~\cite{wei2021groundslam} uses ground submap as a constraint to reduce pose error in terms of roll, pitch, and $z$ error. The ground is extracted based on weighted least-square methods. Then, correspondences between ground at the moment and groud submap are used for pose graph optimization. Accordingly, pose drift caused by LiDAR measurement bias can be reduced, by  minimization of the residual errors from ground constraints. Similar to~\cite{shan2018lego}, Guo \textit{et al.}~\cite{guo20213d} proposed ground-optimized method, yet they employed RANSAC~\cite{fischler1981ransac} for ground plane fitting, which sometimes converge to local minima in urban environments.

\section{PaGO-LOAM: Ground-Optimized\\LiDAR Odometry} \label{sec:m}

\subsection{Overview of Proposed Framework}

The process of the system is based on that of LeGO-LOAM~\cite{shan2018lego}, and it is divided into five steps as follows: 1) Project a point cloud at the moment into the range image, which is followed by ground segmentation based on~\cite{himmelsbach2010fast}. 2) Extract edge features and planar features based on the smoothness~\cite{shan2018lego} in the range image. 3) Estimate the odometry by obtaining the transformation matrix between consecutive scans using the correspondence of the features. 4) Map the features into the global point cloud map. 5) Accumulate a global point cloud map by using the resultant LiDAR odometry. 

In this paper, we enable switching ground segmentation module on the step 1. The following paragraphs highlight the brief introduction of the modified process of ground-optimized LiDAR odometry.

\subsection{Input and Output of Ground Segmentation}

First, the input and output of ground segmentation are briefly introduced. By denoting $\mathbf{z}$ as a 3D point cloud captured by a 3D LiDAR sensor at the moment where $\mathbf{z}$ consists of $N$ points. Then ground segmentation module takes $\mathbf{z}$ and outputs partial cloud points, $\hat{G}$ where, among $\hat{G} \in \mathbf{z}$. Accordingly, the $\mathbf{z}$ is divided into two parts: estimated ground points, $\hat{G}$, and non-ground points, $\hat{G}^c$. That can be expressed as follows: 

\begin{equation}
\mathbf{z} = \hat{G} \cup \hat{G}^c.
\end{equation}

\noindent Next, $\mathbf{z}$ and $\hat{G}$ are projected into the range image and ground image, respectively, which is shown in Fig.~\ref{fig_flow}. In LeGO-LOAM pipeline~\cite{shan2018lego}, projection images are required to perform feature extraction. 
To be more specific, a range image is for feature extraction by calculating smoothness for each pixel in the range image and a ground image is employed as a mask of the image plane to discern planar features from the ground and non-ground objects. Consequently, these features are used for optimization to estimate the relative pose between two consecutive frames.

\subsection{Potential Limitations of Existing Ground Segmentation}


 Originally, LeGO-LOAM uses the angle difference between two points which are located in the same column on the image plane, i.e. a point that corresponds to the $(u, v)$ on the image plane and the other point on the $(u, v-1)$ when extracting the ground. $u$ and $v$ denote two orthogonal coordinate on the image plane. In other words, if the angle difference is less than $\tau_\theta$, two corresponding points are assigned as ground and are not used for edge feature detection~(empirically, $\tau_\theta$ is set to $10^\circ$). 
 
 For these reasons, it tends to be sensitive against some noises because the ground segmentation method only estimates ground points solely based on the geometrical relation of two points. However, in urban environments, the ground can be bumpy as well. Even, some objects such as lawns or bushes can impede the ground segmentation from labeling ground points because their gradients of shape are arbitrary. Accordingly, the ground mask can be mislabeled and these mislabeled parts potentially result in wrong feature correspondence. In the end, it may give rise to imprecise pose estimation.

\subsection{Robust Ground-Optimized LiDAR Odometry}

We substitute the ground segmentation module with other ground segmentation modules to test the effect of ground segmentation on pose estimation. As shown in Fig.~\ref{fig_flow}, the detailed process of the odometry framework is presented. A 3D point cloud captured by a 3D LiDAR sensor is taken as input into the ground segmentation module, e.g. Patchwork~\cite{lim2021patchwork}, and LiDAR odometry framework. Once the ground points are segmented, our method projects a raw point cloud and a ground cloud into two images, respectively. After that, planar and edge feature extraction is performed for the pixel which is not labeled as ground by the ground image. Then, the previously planar features are also extracted in the ground image, followed by the integration of planar features from the ground and non-ground objects. Finally, odometry is estimated by two-stage optimization~\cite{shan2018lego}. 

In summary, robust ground segmentation help perform precise feature extraction. Empirically, our PaGO-LOAM shows better odometry performance compared with baseline method~(see Section~\rom{5}.\textit{C}), which supports that more precise feature extraction leads to accurate ground-optimized odometry.

\section{Experiments} \label{sec:exp}

\subsection{Dataset}

We experimented in the outdoor environments by using SemanticKITTI dataset~\cite{geiger2012kitticvpr, geiger2013vision}. In particular, Seq.~\texttt{00}, Seq.~\texttt{02}, and Seq.~\texttt{05} are used, which consists of 4,531, 4,661, and 2,761 frames, respectively. Note that Seq.~\texttt{00} and Seq.~\texttt{05} are urban scenes, whereas Seq.~\texttt{02} is relatively rural scene~\cite{pan2021mulls}. 

\subsection{Error Metrics}

The experiments are evaluated in terms of 1) ground segmentation and 2) LiDAR odometry by using following metrics.

\subsubsection{For Ground Segmentation}

To evaluate ground segmentation methods quantitatively, \textit{Precision} and \textit{Recall} are employed. Let $N_{\text{TP}}$, $N_{\text{TN}}$, $N_{\text{FP}}$, and $N_{\text{FN}}$ be the number of points in TP, TN, FP, and FN, \textcolor{black}{respectively}; then, precision and recalls are defined as follows:
\begin{itemize}
\item Precision: $\frac{N_{\text{TP}}}{N_{\text{TP}}+N_{\text{FP}}}$, \; Recall: $\frac{N_{\text{TP}}}{N_{\text{TP}}+N_{\text{FN}}}$.
\end{itemize}

\subsubsection{For LiDAR Odometry}

On the other hand, to evaluate performance of odometry,  the relative odometry errors,  $t_{\text{rel}}$ for relative translation error and $r_{\text{rel}}$ for relative rotation error, are used as quantitative metrics.
In addition, absolute trajectory error (ATE) is also used to compare total errors of trajectories. The three metrics are calculated by \cite{Zhang18rpg_eval_tool}.

\section{Results and Discussion} \label{sec:resuls_and_discussion}

\subsection{Qualitative Analysis of Ground Segmentation}

\begin{figure}[t!]
\centering
    \captionsetup[subfigure]{justification=centering}
    \subfloat[\centering Ground segmentation module in LeGO-LOAM~\cite{shan2018lego}]{
    \includegraphics[width=0.23\textwidth]{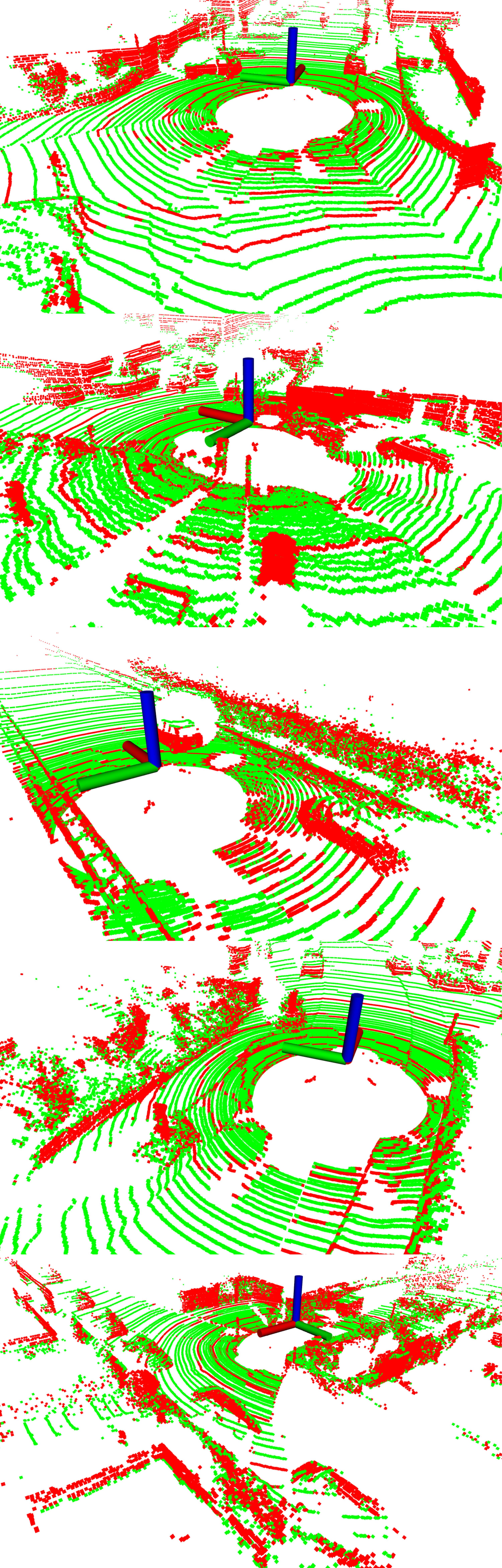}%
    }
    \captionsetup[subfigure]{justification=centering}
    \subfloat[Patchwork~\cite{lim2021patchwork}]{
    \includegraphics[width=0.23\textwidth]{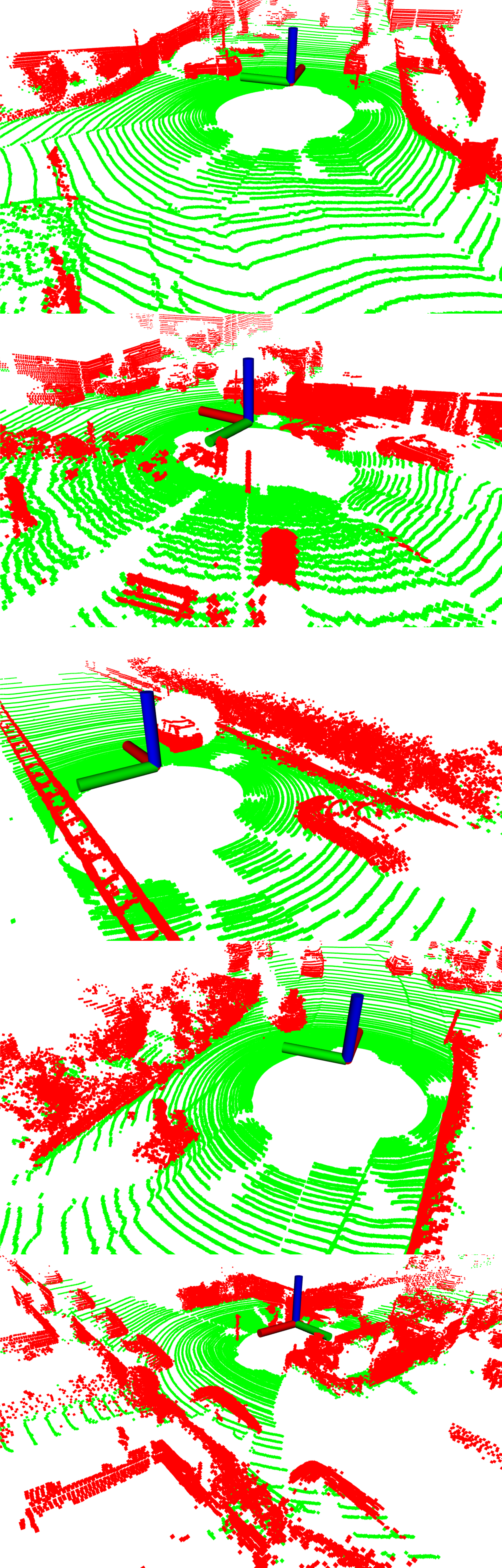}%
    }
    
    \captionsetup{font=footnotesize}
    \caption{Qualitative comparison between ground segmentation module implemented in LeGO-LOAM~\cite{shan2018lego} and Patchwork~\cite{lim2021patchwork}. It was shown that the former method is sensitive when encountering undulated terrains or bushy regions, thus it results in many false negatives. On the other hand, Patchwork~\cite{lim2021patchwork} shows robust ground segmentation performance. The red and green colors denote estimated non-ground points and ground points, respectively (best viewed in color).}
    \label{fig:comparison_ground_seg}
\end{figure}

First of all, a qualitative comparison between the ground segmentation module in LeGO-LOAM~\cite{shan2018lego} and Patchwork~\cite{lim2021patchwork} was conducted. As shown in Fig.~\ref{fig:comparison_ground_seg}, the ground segmentation within LeGO-LOAM showed detailed ground estimations. However, it struggles with nonflat areas, including a steep slope, a complex intersection, and a region where many curbs exist. In particular, the module is likely to be sensitive when encountering bumpy terrains or bushy regions, which is the potential limitation of line-based ground segmentation~\cite{narksri2018slope}. For these reasons, it gives rise to under-segmentation, which is represented as red lines in the ground. 

 In contrast, Patchwork~\cite{lim2021patchwork}, which is the SOTA ground segmentation method, shows relatively smooth and continuous ground segmentation performance with a few false negatives. Unlike the ground segmentation module in LeGO-LOAM, Patchwork estimates ground based on the plane fitting for each unit space called bin. Consequently, thus it is more robust in uneven environments. 
 
 In particular, Patchwork~\cite{lim2021patchwork} is based on the premise that ground points are located in the lowest parts along the $Z$ direction within a bin. It is remarkable that it outputs few false positives. In contrast, the ground segmentation within LeGO-LOAM sometimes considers upper parts of bushes or cars as ground points. This phenomenon may lead to imprecise feature matching in odometry frameworks.

\subsection{Quantitative Analysis of Ground Segmentation}

\begin{figure*}[t!]
    \captionsetup{font=footnotesize}
	\centering 
	\includegraphics[width=1.0\textwidth]{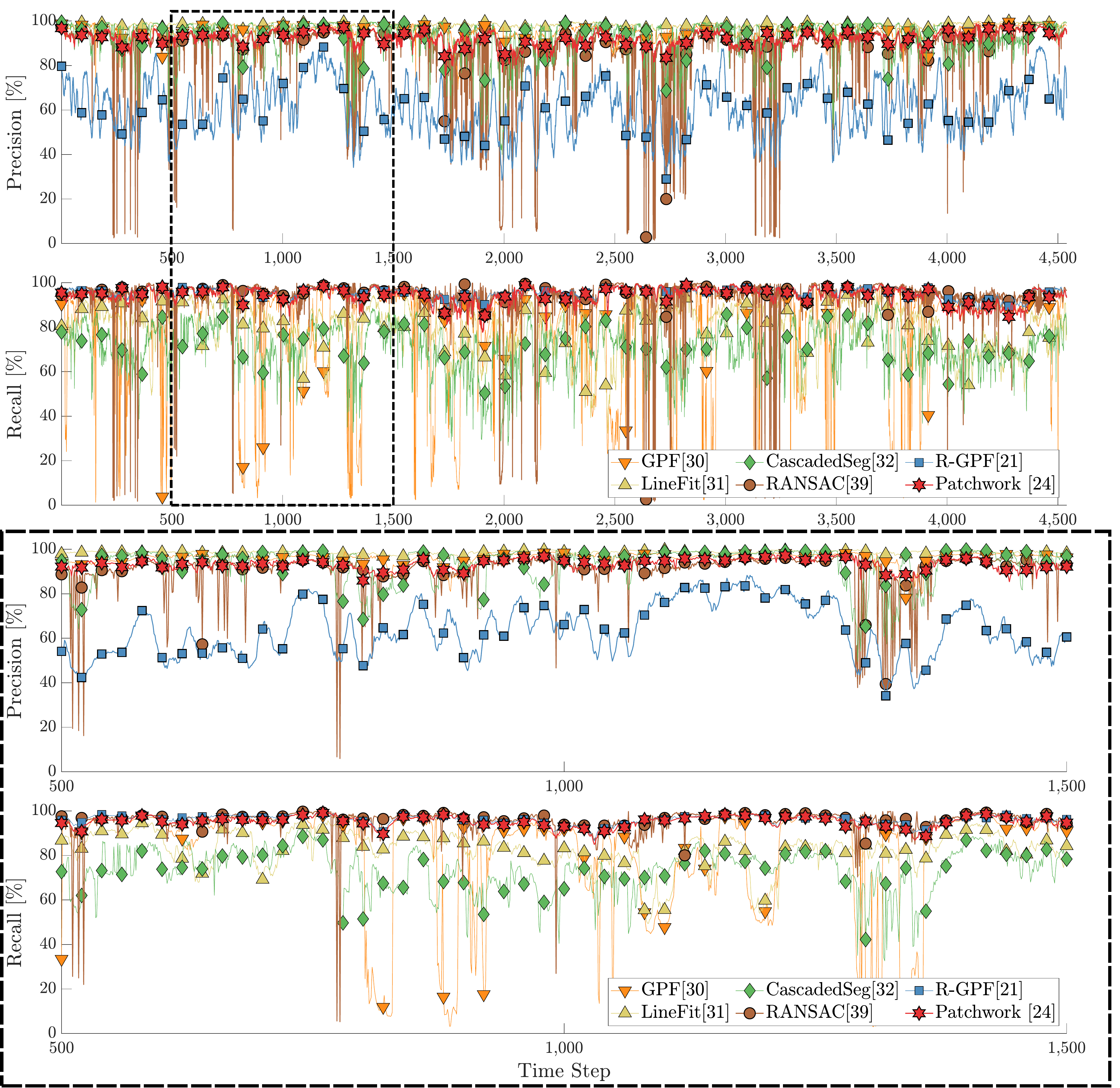}
	\caption{Precision and recall on Seq.~\texttt{00} in the SemanticKITTI dataset~\cite{behley2019semantickitti}. A ground segmentation method should guarantee consistency for stable feature extraction and matching in LiDAR odometry. Thus, the less fluctuating the performance, the better. In this sense, Patchwork~\cite{lim2021patchwork} shows a promising performance, with little variance of recall relative to other methods.}
	\label{fig:sequential_comparison}
	\vspace{-0.7cm}
\end{figure*}

Next, quantitative analysis was conducted to compare the SOTA ground segmentation methods, namely, R-GPF~\cite{lim21erasor}\footnote{https://github.com/LimHyungTae/ERASOR}, Patchwork~\cite{lim2021patchwork}\footnote{https://github.com/LimHyungTae/patchwork}, Line-Fit\footnote{https://github.com/lorenwel/linefit\_ground\_segmentation}~\cite{himmelsbach2010fast}, RANSAC \cite{fischler1981ransac}, GPF\footnote{https://github.com/VincentCheungM/Run\_based\_segmentation} \cite{zermas2017fast},  CascadedSeg\footnote{https://github.com/n-patiphon/cascaded\_ground\_seg} \cite{narksri2018slope}. Note that performance of ground segmentation should not be fluctuated because LiDAR odometry takes sequential data as input. In other words, ground segmentation should guarantee consistency for stable feature extraction and matching. Thus, the less fluctuating the performance, the better.

In this sense, Patchwork shows a promising performance, as shown in Fig.~\ref{fig:sequential_comparison}. In particular, Patchwork estimates the ground with little variance of recall relative to other methods. This confirms that our method tackle the under-segmentation problem, while estimating ground segmentation consistently. 

 On the other hand, it was shown that other methods sometimes fail to estimate ground segmentation, as presented in Fig.~\ref{fig:sequential_comparison}. This indicates that other methods sometimes converge to a local minimum. For instance, RANSAC shows both low precision and low recall once in a while because there exist dominant planes compared with the ground in urban environments, such as walls. On the other hand, GPF~\cite{zermas2017fast} and CascadedSeg~\cite{narksri2018slope} are region-wise methods like Patchwork, but the way to divide regions is somewhat na\"ive so the size of bins are too big; thus, it is not safe to assume that the ground is planar within the bin. Consequently, the assumption does not hold, resulting in under segmentation.
 
 Therefore, both qualitative and quantitative analyses address that Patchwork is the most proper ground segmentation method as preprocessing in complex urban environments.

\subsection{Effect of Ground Segmentation on LiDAR Odometry}

Finally, the effect of ground segmentation on odometry is checked. We compare LeGO-LOAM~\cite{shan2018lego} and the LiDAR odometry whose ground segmentation module is replaced with Pacthwork, called PaGO-LOAM. The methods are evaluated for comparison in a)~no-loop-closing situations and b)~loop-closing situations, respectively. 

\begin{figure*}[thp]
\centering
    \captionsetup[subfigure]{justification=centering}
    \subfloat[LeGO-LOAM~\cite{shan2018lego}]{
    \includegraphics[width=0.45\textwidth]{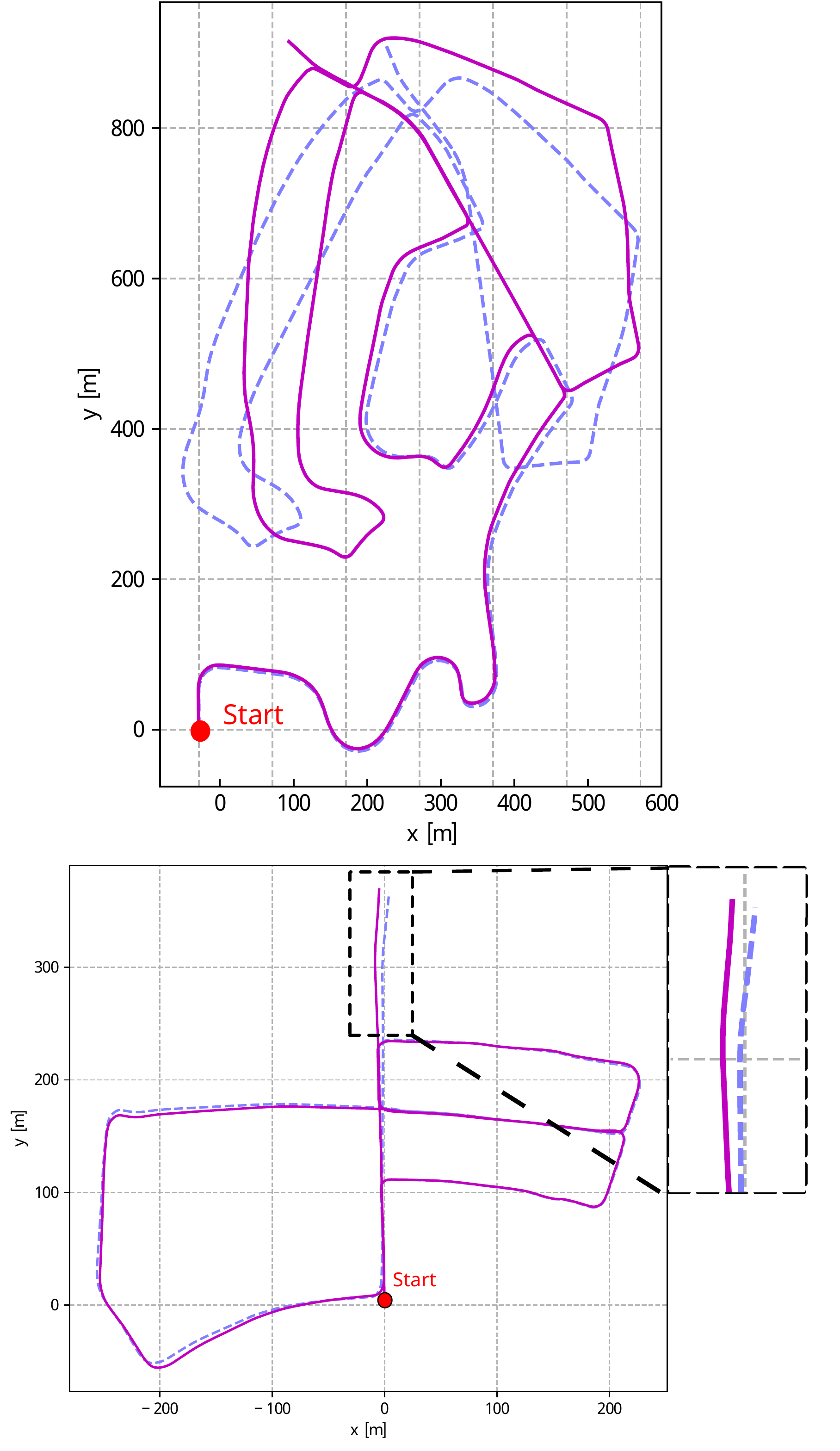}%
    }
    \captionsetup[subfigure]{justification=centering}
    \subfloat[PaGO-LOAM~(Ours)]{
    \includegraphics[width=0.45\textwidth]{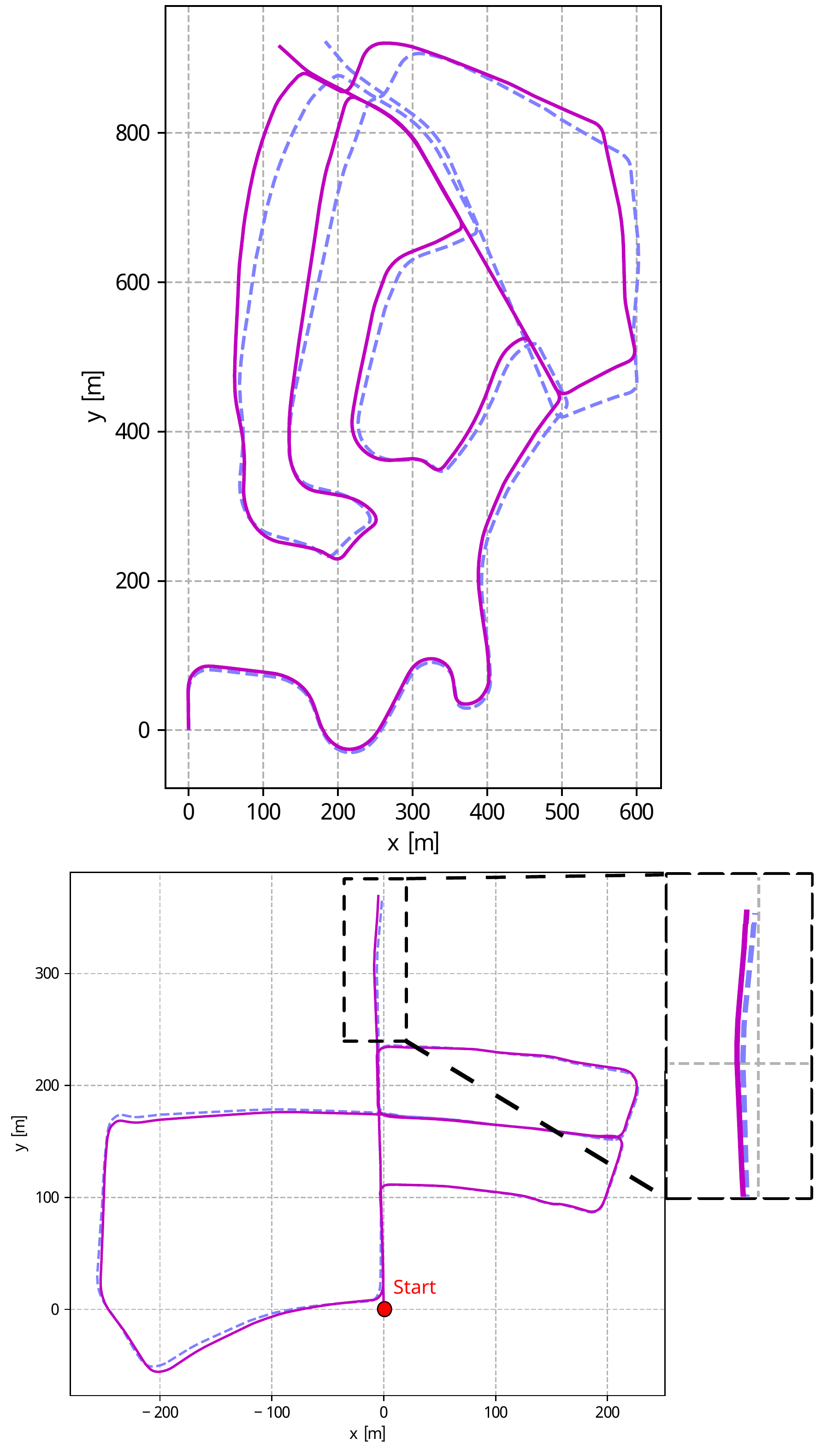}%
    }
    \caption{Comparison between our proposed method and LeGO-LOAM~\cite{shan2018lego}. (T-B): Trajectories on Seq.~\texttt{02} and Seq.~\texttt{05} in the KITTI dataset~\cite{geiger2012kitticvpr}. The dashed line denotes estimated trajectory and the solid line denotes the ground truth.}
    \label{fig:kitti_traj}
    \vspace{-0.3cm}
\end{figure*}


\begingroup
\begin{table}[th]
	\centering
	\caption{Comparison of relative pose error (RPE) for our proposed method with LeGO-LOAM on Seq.~\texttt{00}, Seq.~\texttt{02}, and Seq.~\texttt{05} of the KITTI dataset~\cite{geiger2012kitticvpr}. LC is an abbreviation for loop closing. All the metrics are the less, the better ($t_{\text{rel}}$: \%, $r_{\text{rel}}$: deg/100m).}
	\setlength{\tabcolsep}{4pt}
	{\scriptsize
	
	\begin{tabular}{llcccccc}
	
	\toprule \midrule
	&\multirow{2}[3]{*}{Method} & \multicolumn{2}{c}{\texttt{00}} & \multicolumn{2}{c}{\texttt{02}} & \multicolumn{2}{c}{\texttt{05}} \\  \cmidrule(lr){3-4} \cmidrule(lr){5-6} \cmidrule(lr){7-8} 
	&  & $t_{\text{rel}}$ & $r_{\text{rel}}$ & $t_{\text{rel}}$ & $r_{\text{rel}}$ & $t_{\text{rel}}$  & $r_{\text{rel}}$   \\ \midrule
	\multirow{2}{*}{W/o LC} &LeGO-LOAM ~\cite{shan2018lego} & 0.90 & 0.30 & 2.71 & 0.72 & 0.79 & \textbf{0.31} \\
	& PaGO-LOAM (Ours) & \textbf{0.87} & \textbf{0.30} & \textbf{1.31} & \textbf{0.35} & \textbf{0.76} & 0.32 \\ \midrule 
	\multirow{2}{*}{W/ LC} &LeGO-LOAM ~\cite{shan2018lego} & 0.86 & \textbf{0.24} & 2.97 & 0.73 & 0.82 & 0.34 \\
	 & PaGO-LOAM (Ours) & \textbf{0.85} & \textbf{0.24} & \textbf{1.33} & \textbf{0.38} & \textbf{0.81} & \textbf{0.33} \\ 
	\midrule \bottomrule
	\end{tabular}
	}
	\label{table:kitti_rpe}
	\vspace{-0.3cm}
\end{table}
\endgroup

\begin{table}[th]
	\centering
	\caption{Comparison between our algorithm and LeGO-LOAM in terms of Absolute Trajectory Error (ATE) on Seq.~\texttt{00}, Seq.~\texttt{02}, and Seq.~\texttt{05} of the KITTI dataset~\cite{geiger2012kitticvpr}. LC is an abbreviation for loop closing. The less, the better (unit: m).}
	
	{\scriptsize
	
	\begin{tabular}{llccc}
	\toprule \midrule
	 & Method & \texttt{00} & \texttt{02} & \texttt{05} \\ \midrule
\multirow{2}{*}{W/o LC} & LeGO-LOAM~\cite{shan2018lego} & \hfil 6.01 & \hfil 57.24 & \hfil 2.22\\
 & PaGO-LOAM (Ours) & \hfil \textbf{5.03} & \hfil \textbf{13.45} & \hfil \textbf{1.55}\\ \midrule
        \multirow{2}{*}{W/ LC} & LeGO-LOAM~\cite{shan2018lego} &\hfil \textbf{2.31} &\hfil 57.83 &  \hfil 1.84\\
& PaGO-LOAM (Ours) & \hfil 2.38 & \hfil \textbf{12.95} & \hfil \textbf{1.60} \\
	\midrule \bottomrule
	\end{tabular}
	}
	\label{table:kitti_rmse}
	\vspace{-0.4cm}
\end{table}

As a result, our proposed LiDAR odometry shows a better performance compared with that of LeGO-LOAM as summarized in Table~\ref{table:kitti_rpe} and Table~\ref{table:kitti_rmse}. In particular, as also shown in Fig.~\ref{fig:kitti_traj}, there are remarkable differences in Seq.~\texttt{02}. As mentioned in Section~\rom{4}.\textit{A}, Seq.~\texttt{02} is a rural scene, so the ground is more uneven and has a steep slope. For this reason, the original ground segmentation module in LeGO-LOAM may occasionally fail to estimate the ground properly. In other sequences, there were no recognizable differences in performance, so we just made a table only on Seq.~\texttt{00}, Seq.~\texttt{02} and Seq.~\texttt{05}

Therefore, we conclude that precise and robust ground segmentation could increase the performance of the LiDAR odometry, particularly in rural scenes where the ground becomes uneven and has steep slopes.

\section{CONCLUSION} \label{sec:con}

In this study, a robust ground-optimized LiDAR odometry framework has been proposed. In particular, \textit{PaGO-LOAM} has been proposed by leveraging the SOTA ground segmentation method called Patchwork~\cite{lim2021patchwork}. Our framework allows easy accessability to check the effect of ground segmentation on the odometry performance. In future works, we plan to propose more robust ground-optimized odometry method, as well as conduct close quantitative evaluations.

\bibliographystyle{IEEEtran}
\bibliography{./ur2022,./IEEEabrv}

\end{document}